# AUTOMATIC ESTIMATION OF LIVE COFFEE LEAF INFECTION BASED ON IMAGE PROCESSING TECHNIQUES

Eric Hitimana[*1] and Oubong Gwun[*2]

[*]Department of Computer Science and Engineering, Chonbuk National University,
Jeonju City, South Korea
[1]hitimeric06@yahoo.fr, [2]obgwun@jbnu.ac.kr

## ABSTRACT

*Image segmentation is the most challenging issue in computer vision applications. And most difficulties for crops management in agriculture are the lack of appropriate methods for detecting the leaf damage for pests' treatment. In this paper we proposed an automatic method for leaf damage detection and severity estimation of coffee leaf by avoiding defoliation. After enhancing the contrast of the original image using LUT based gamma correction, the image is processed to remove the background, and the output leaf is clustered using Fuzzy c-means segmentation in V channel of YUV color space to maximize all leaf damage detection, and finally, the severity of leaf is estimated in terms of ratio for leaf pixel distribution between the normal and the detected leaf damage.*

*The results in each proposed method was compared to the current researches and the accuracy is obvious either in the background removal or damage detection.*

## 1. INTRODUCTION

A computer vision system is an attempt to replicate the human eye to brain assessment process, whereby the human eye is replaced by a digital camera and the human brain is replaced by a learning algorithm. The camera can record objective and consistent image data without substantial confounding noise [1].

Image processing has been proved to be effective tool for analysis in various fields and applications [2]. In evolution towards sustainable agriculture system it was clear that important contributions can be made by using emerging techniques. Precision agriculture was new and developing technology which leads to incorporate the advance techniques to enhance farm output and also enrich the farm inputs in profitable and environmentally sensible manner. With these techniques/ tools it was now possible to reduce errors, costs to achieve ecological and economical sustainable agriculture.

Coffee rust is the most economically important coffee disease in the world, and in monetary value, coffee is the most important agricultural product in international trade. Even a small reduction in coffee yields or a modest increase in production costs caused by the rust has a huge impact on the coffee producers, the support services, and even the banking systems in those countries whose economies are absolutely dependent on coffee export.

Infections occur on the coffee leaves. The first observable symptoms are small, pole yellow spots on the upper surfaces of the leaves. As these spots gradually increase in diameter, masses of orange urediniospores (or uredospores) appear on the under surfaces. The fungus sporulates through the stomata rather than breaking through the epidermis as most rusts do, so it does not form the pustules typical of many rusts. The powdery lesions on the undersides of the leaves can be orange-yellow to red-yellow to red-orange in color, and there is considerable variation from one region to another [3].

Coffee rust and other coffee pests cause premature defoliation, which reduces photosynthesis capacity and weakens the tree [4]. The most techniques used to avoid coffee rust and other pests that destroy the coffee leaves are pesticides and fungicides, but if they are not controlled well, they can cause ecosystem problems.

The detection of severity of infected leaves have been done by the farmers using naked eyes, which can contribute to many errors, and the precise ways are needed to be sure the amount of pesticides or fungicides to be applied while preserving ecosystem. Most leaf diseases destroy the leaf, so that it can be easy to detect the damage using image processing techniques, but in the case of coffee rust, there is only color change which acts as a special case.

From the shown problems, we propose a method for detecting the pests' attacks infection on image of coffee's leaf using image segmentation techniques.
In this paper, the novelty is that the images used are captured from the tree, to avoid defoliation. Most researches about leaf disease detection [5], [6], [7], have been done, but they cut off the leaf and put it to the white background for easy processing, but our algorithm considers all leaves' images regardless the type of background.

As far as the system is concerned, our algorithm is made by three processes. Firstly, the captured coffee leaf is processed for contrast enhancement using LUT gamma correction algorithm by Sayaraman et al. [17]. Secondly, the enhanced image is processed to remove the unwanted background. At this stage, the captured image maybe having many surrounding leaves or branches of trees, the concern is to detect the main leaf to be processed separately. Lastly, the recovered leaf is segmented using fuzzy C mean clustering to detect the infected part of the leaf; the severity of infected leaf area is estimated by calculating the ratio of the infected pixel distributions to the normal leaf pixel distributions.

## 2. RELATED WORKS

In this section, we survey the related current researches on image processing in agricultures, such as background removal, infected leaf segmentation and area of leaf measurement.

### 2.1. Leaf disease detection and classification

As a rapid, nondestructive and objective method, image processing technology has been widely used in determination of some quality characteristics of agricultural products. Leaf disease detection and classification is a hot research in plant managements and taxonomy.

Gloria D. et al. [5] proposed semi-automatic approach based on an initial pixel classification according to the chrominance feature from the YCrCb color space. It requires user-intervention to select a sample of pixels for training the color space classifiers. Thiago L. G. Souza et al. [6] automatically classified the main agents that cause damages to soybean leaflets. After extracting the contour of the damages, they are taken as a complex network, and trained using SVM.

C.P. Wijekoon, et al. [9] used Scion Image software to quantify a wide variety of fungal interactions with plant leaves. This software is responsible for measuring the change in leaf color caused by fungal sporulation or tissue damage. But it only deals with the detached, well placed and shadow free leaves. Qinghai He et al. [10] proposed the damage detection method by measuring the damage ratio in different color model after enhancing the leaf image, but the algorithm fails to handle the outdoor leaves for contrast and other noises. A.C. Nazare-JR. et al. [11] automatically quantified the damaged leaf area by handling the noises and recovering the leaf contour using computational geometry, but they recovered only the line segments but not the curved edges and they only detected the damaged/ destroyed parts, not the damage part in terms of color change, as we are considering the healthy leaf as the one responsible for photosynthesis process.

All of those methods only handle the detached, well placed, shadow free, with simple background leaves images, and thus they only did the simple image segmentation (thresholding) to remove the background noises. On the other hand, our proposed algorithm considers all types of leaves regardless any background, and can also detect all damaged leaf parts (in terms of destroying and color change).

### 2.2. Background removal

In still image object detection, many researchers proposed different algorithms for background removal for image segmentation purpose. Jeong-In Park et al. [18] suggested the variable order n x m dimensional vector, where the vectors are applied to the reduced objective image to remove the background. This method does not remove the actual background; rather it regenerates the filtered replica after replicating the background. It may look like the background is actually removed after it is applied to the image, but it is reconfigured with white color lines which are smoothly processed while retaining the background. This method have been proposed to overcome the computational time of code book, but compare to our method, this is still expensive in terms of computation depending on the size of objective image as it is dealing with image reduction and n x m dimensional vector processing.

In object extraction based on detecting salient regions [19], [20], [21], [22] they ended by segmenting the image to remove the background, but sometimes this method failed to detect the object in question. Guanqun Cao et al. [22] proposed a salient object extraction with opponent color boosting, the method is based on emphasis on color in an iso-salient color space and filtering by a DoG filter afterwards, but the detection is not accurate as it includes even other non objects. Our method has been compared with this one for object detection.

In most agriculture image processing applications, they applied a simple threshold or otsu's image segmentation [23] [11], [6] to separate the background and the foreground because the background is not complicated. Our proposed method for background removal provides an obvious accuracy to their papers as well.

### 2.3. Leaf Area Measurement

Accurate and rapid non-destructive leaf area measurement/estimation is important in plant understanding and modeling ecosystem function. Utilizing the leaf area instruments, it is reliable and convenient to estimate leaf area using mechanical, digital or portable scanning planimeters [13], but the method is expensive and destructive i.e., means we need to cut off the leaf or some tools can not fit the whole leaf, so it need to be cut into different pieces and measure each piece separately and add together after.

Mahdi M. Ali et al. [14], after preprocessing the leaf and detecting the edge, they used digital vernier and compared the results with Li-Cor 3100 leaf area meter. The method is not destructive but still they used some devices to measure the area. Chaohui Lu et al. [8] captured the leaf and put on a hand normal panel, i.e. square (known area) drawn on a white paper, the image is processed using image processing techniques, and then the area is calculated through pixel number statistic. The accuracy can be affected by the geometric distortion of the panel. Sanjay B. Patil et al. [15], after processing the leaf image, the area is calculated by estimating the pixel statistic referencing to piece of coin (known area). The pixel count of the processed image depends on the distance between the camera and the object when the picture is taken.

In this paper, the proposed algorithm handles all leaf images regardless where and how the image was taken (background, environmental condition that can be affecting the contrast of image, etc), we ended by finding that to adopt one of the proposed method above, can lead us to many errors, because we did not care how far or close the image leaf were captured/shot.

To overcome those challenging, we decided to estimate the severity of the damaged area in percentage by calculating the ratio between the normal leaf pixel distributions (statistic) to the infected pixel distributions.

## 3. PROPOSED METHOD

This section describes the process of our proposed method step by step. Our algorithm consists of three steps: Image contrast enhancement, background removal and detection of the damaged area with its severity estimation. The figure 1 shows the overview of the system.

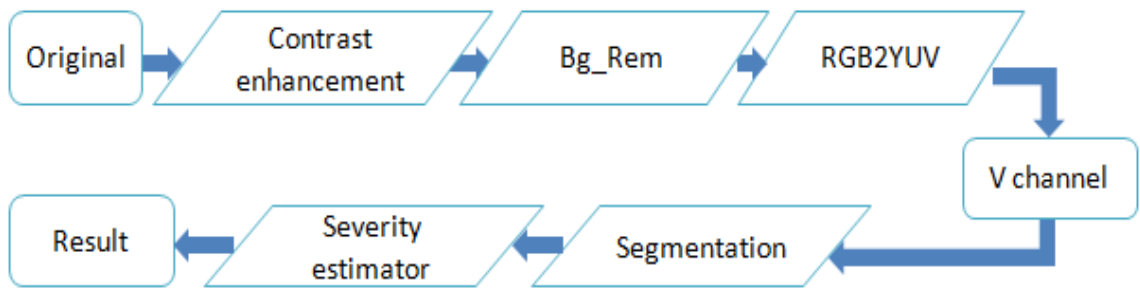

Figure 1: Framework of the proposed method

### 3.1. Contrast enhancement

The contrast enhancement process adjusts the relative brightness and darkness of objects in the scene to improve their visibility. As our method can handle any image regardless of its condition of shot, we decided to use LUT based gamma correction algorithm to deal with the image contrast enhancement.

$$LUT = Max\_int \; x \left( \frac{0:Max\_int}{Max\_int} \right)^{1/\gamma} \quad (1)$$

$$I' = LUT(double(I)) \quad (2)$$

Look-Up Table (LUT) is formulated using the maximum intensity (Max_int, equals to 255), and the value of gamma depends on the input image.

The original brightness value of image I is mapped to I' by using the formulated LUT as shown on figure 2.

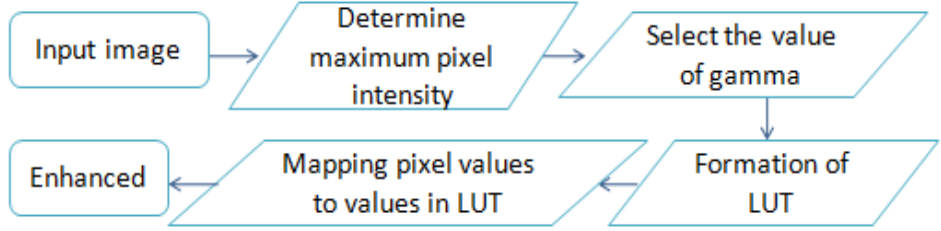

Figure 2: Image enhancement using gamma correction

We proposed a method to set the value of gamma automatically based on the characteristics of the original image.

1. Analyze the histogram of an input gray image,
2. Calculate the mean average intensity value

$$I_{avg} = \frac{1}{WxH} \sum_{i=1}^{W} \sum_{j=1}^{H} I(i,j) \quad (3)$$

3. Normalize the value in range [0,1]

$$r = \frac{I_{avg}}{I_{max}} \quad (4)$$

4. Get the gamma value by using equation

$$\gamma = \begin{cases} 10r - 4 & \text{for } r>0.5 \\ 1 & \text{for } r=0.5 \\ r^{0.1} - 0.4 & \text{for } r<0.5 \end{cases} \quad (5)$$

In our experimental results, there is no doubt to say that in both contrast conditions (low and high); the gamma value is selected efficiently. And this proposed automatic gamma value always enhances the image based on its average gray intensity value, this method is different from the adaptive gamma correction proposed by Shi-Chia Huang et al. [12] which requires some adjustable parameters and it always enhances the low contrast, because the value of gamma is always equal or less than 1 according to their equation $\gamma = 1 - cdf_w(l)$. The figure 3 and 4 show our results for contrast enhancement.

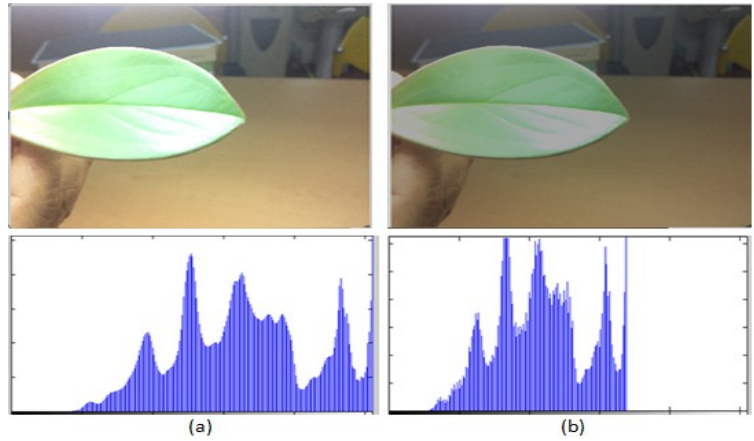

Figure 3: Result for contrast enhancement images and their gray value histograms: (a) original image of high contrast, (b) resulting image with adjusted contrast for better processing.

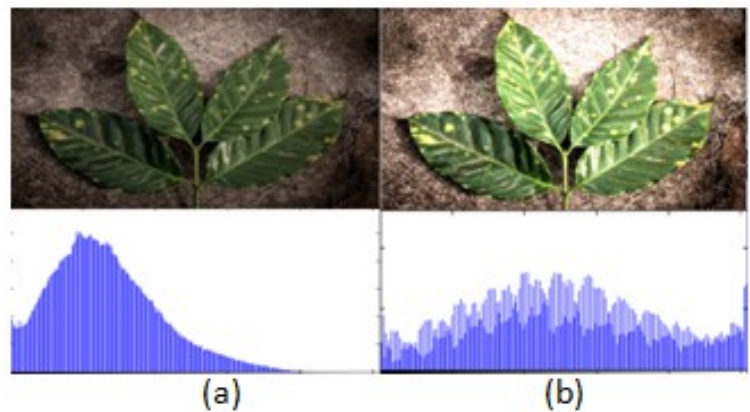

Figure 4: Result for contrast enhancement, images with their gray value histograms: (a) original image of low contrast, (b) resulting image with enhanced contrast for better processing.

### 3.2. Background removal

The leaf image processed maybe having some surrounding noises that can affect the accuracy of leaf damage detection. We decided to propose a method that can only keep the foreground object, i.e., leaf only.

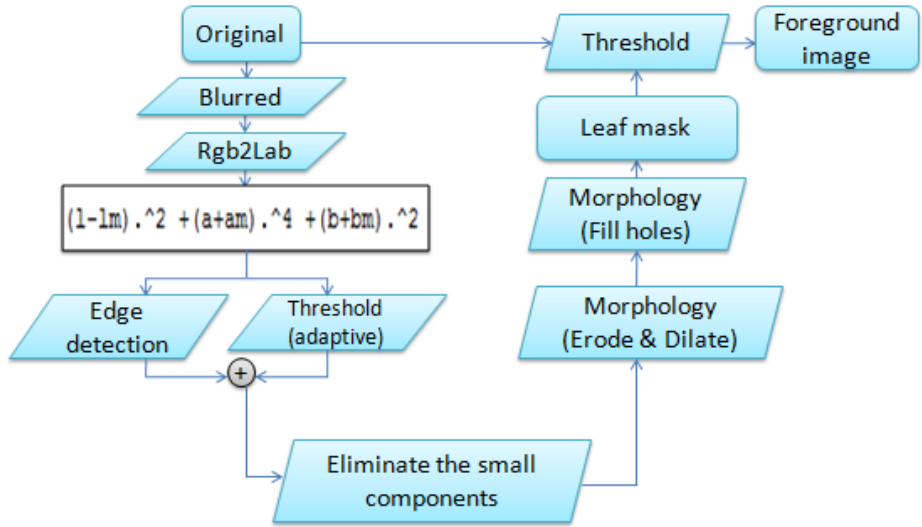

Figure 5: Proposed method for background removal

The original image is blurred using Gaussian kernel to suppress the noise, and the image is converted to CIELab color space that have been proved as the most color space to detect the object based on the salient properties [17].

$$O(x, y) = (I_{\mu l} - \tilde{I}_l)^2 + (I_{\mu a} - \tilde{I}_a)^4 + (I_{\mu b} - \tilde{I}_b)^2 \quad (6)$$

Where the $I_\mu$ is the arithmetic mean value of the image in each channel, $\tilde{I}$ is the corresponding image pixel vector value in the Gaussian blurred version (using a 5x5 separate binomial kernel) of the original image. The above proposed equation (6) can highlight the foreground object and suppress the background. The resulting image is threshold using the following adaptive threshold.

$$Th_r = \frac{1}{WxH} \sum_{i=1}^{W} \sum_{j=1}^{H} O(x, y) \quad (7)$$

The threshold image is combined by the boundary features detected using canny edge detector to adjust the overall structure of the object. The resulting image with different separate objects is judged to remains with the biggest object among the arrays using labeling method. Image erosion and dilation algorithms are applied to adjust the objects by using the disk element of fixed size. The output image is filled to recover the internal holes. The final image is a mask of the whole object within an image, and it is used as a threshold to segment the original image. The resulting output image at this stage is the image with background free as shown in the figure 6 below.

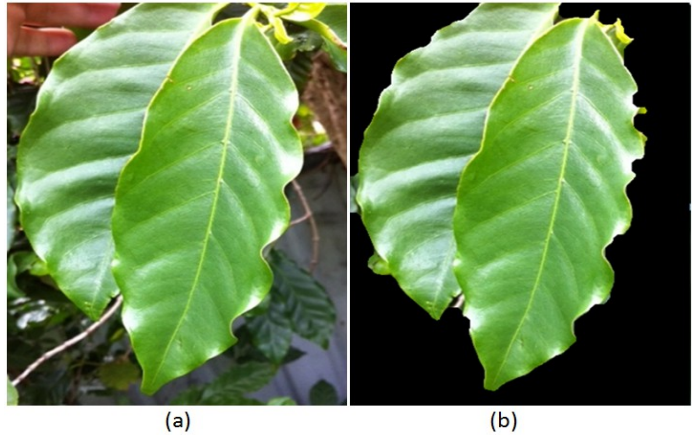

Figure 6: Result for background removal: (a) original image, (b) background free image

### 3.3. Damage clustering

At this stage, the leaf in question is available; the only problem is to detect the damage. In most proposed methods [11], [6], they tried to detect damage in gray image because their method only captured the destroyed part of the leaf as an infected leaf part. But our method considers a damaged part as all leaf areas that cannot contribute to the photosynthesis process.
In our case, we detected the damage leaf area in YUV color model, and our algorithm shows a good efficiency compared to other methods. And the other advantage for using YUV color model is that, the leaf veins are not mistaken as the damage. The V channel is clustered using Fuzzy C-Mean algorithm, where we only used two clusters.

### 3.4. Damage estimator

In this paper, we decided to estimate the severity of the leaf damage, to allow the farmers being able to take into account their plant management (for pesticide or fungicides utilization).

After surveying different methods used for estimating leaf damages, we can say that most of them cannot give good results on our samples image database, we decided to estimate infection by calculating the percentage of the damaged pixels statistic to the normal real leaf pixels distribution.

$$l_{severity} = \frac{l_{\inf ected}}{l_{normal}} x100 \qquad (8)$$

### 4. EXPERIMENTAL RESULTS

In order to validate and test our proposed method, we tested it to many type of leaves and have good results. We compared the results with other researches that have been done. Figure 7 shows the comparison of the proposed method for image background removal. Our proposed method was specific for leaf images, but it can also work better than other methods used for object detection.

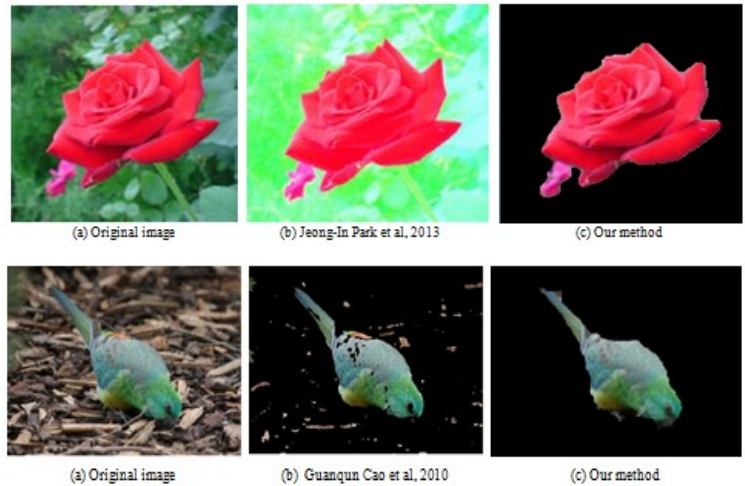

Figure 7: Comparative results of our background removal method.

The figure 8 describes the comparison for our infected leaf detection. Our method for detecting damaged parts of leaf; it cares all part of leaf that cannot contribute to the photosynthesis process which is the main function of the leaf on the plant. Whereas the method of Nazare A.C [11], only considered the damaged leaf as the destroyed one, which is a wrong perception in terms of anatomical process of plants.

As we can see the last row in figure 8, method of Nazare took the tested leaf as a healthy leaf, and according to anatomical concept of the plant, our method can come up with accurate leaf disease detection with estimation of 26,25%.

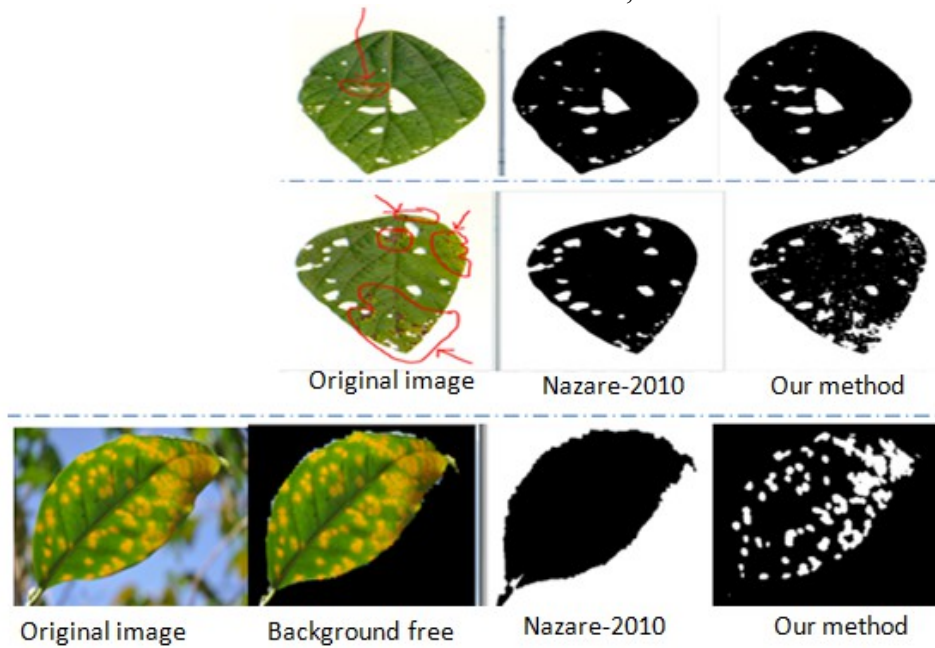

Figure 8: Infected damage detection comparison

The proposed method was applied to a big image database, and the detection was accurate and efficient. Figure 9 and 10 show more results, the former provides the estimate of 7%, and the latter shows the damage at 22%. This is an estimate, because we can see in figure 10 that the trunk was mistaken as an infection.

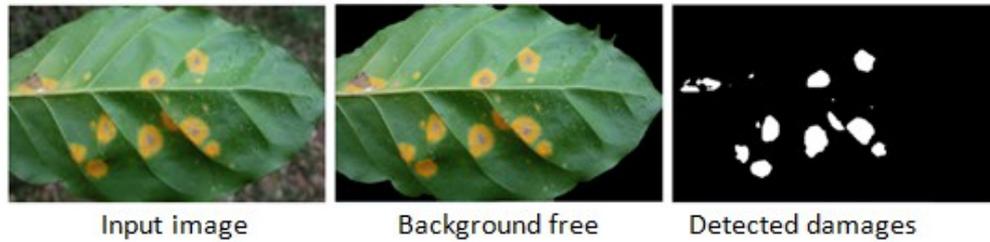

Figure 9: Detection with an estimation of 7%

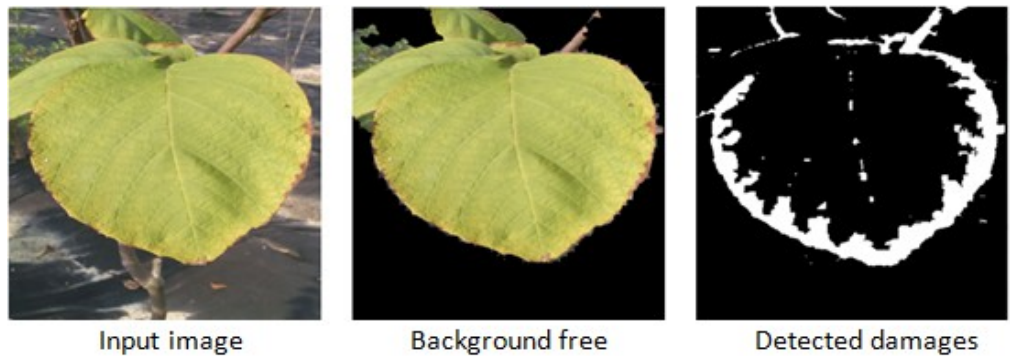

Figure 10: Detection with an estimation of 22%

Nazare et al. evaluated their method by comparing with the manual segmented data by the expert in the area of Plant Science, and other proposed method of (Mura). And according to their results at that time, their method was better from others.

The diagram below shows how accurate our method is compared by Nazare's method. From the same 27 tested images, our average detection was two times than Nazare's method ( $12.36 \pm 8.43$ and $5.76 \pm 4.65$ respectively, i.e. $\mu \pm \delta$ for mean and standard deviation). The strong point for our method is that we can handle all leaf damages (destroyed and color change), it can be seen from the graph that for the last six leaves, which are infected by coffee rust as described in the introduction, Nazare's method considered them healthy, while are damaged already.

On other hand, it is also obvious that for the same destroyed leaf damages, the estimated values are almost the same, which shows that, our method works like their method and beyond for color change leaf damages.

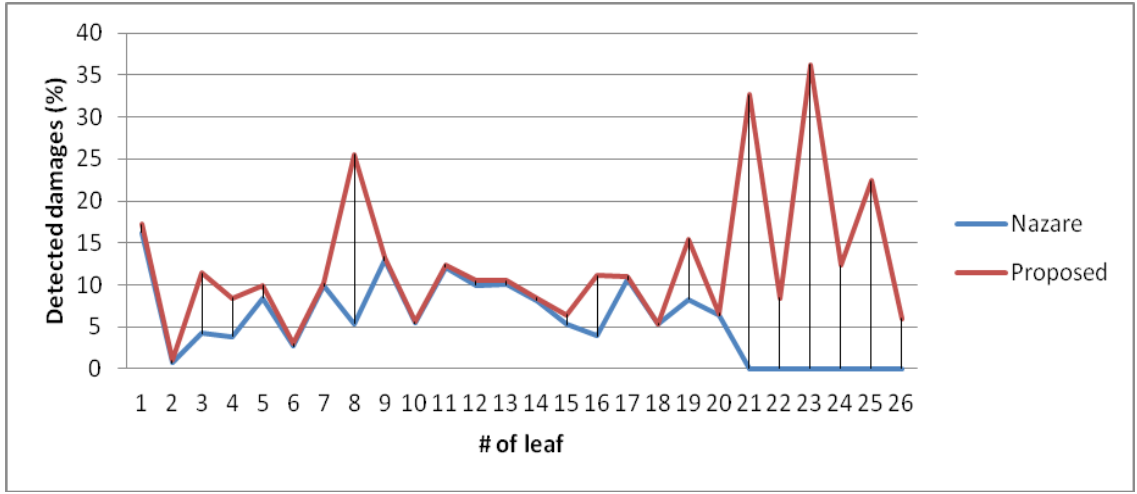

Figure 11: Comparative representation of our method and Nazare's method

## 5. CONCLUSION

In this paper, we proposed an automatic infected leaf detection algorithm that combines three processes: Image contrast enhancement, Image background removal, and estimation of detected infection. After adjusting the contrast by getting the value of gamma automatically, the system processes the original leaf image to keep the real leaf (foreground) by using the background removal method which is based on luminance and color. The background free image is then processed in YUV color model, i.e. on V channel, to maximize the detection of the leaf damage using the Fuzzy C-means Clustering.

The estimation of the severity of infected leaf was fast and quantitatively maximized all leaf damages compared to other methods and the necked eye process used by the farmers. It can help farmers to be sure which quantity of pesticides or fungicides their fields (coffee) require.

The proposed method was compared with some current researches, and it is obvious that it can over perform them either in background removal or in infection detection, even the method is fast; and avoids the defoliation done by all other methods surveyed. In the future, we are planning to upgrade our algorithm in real-time approach.

# REFERENCES

[1] Hasan M. Velioglu and Ozgur Sanglam, 'Evaluation of insect Damage on Beans using Image Processing Technology', 2012.

[2] Anup Vibhute and S K Bodhe 'Applications of Image processing in Agriculture: A Survey', International Journal of Computer Applications Volume 52- No 2, 2012.

[3] Robert H. Fulton, Richard A. Frederiksen, 'Coffee Rust in the Americas', The American Phytopathological Society, St. Paul, Minnesota, 1984

[4] Tropical Plant Diseases by Thurston, H.D, 1998. American phytopathological Society, St. Paul, Minnesota, p123-127

[5] Gloria Diaz, Eduardo R. Juan R. B. Norberto M., Recognition and Quantification of Area Damaged by Oligonychus Perseae in Avocado Leaves, 2009

[6] Thiago L., G,Souza, Eduardo S. M., Kayran Dos S, David M., Application of Complex Networks for automatic classification of damaging agents in Soybean Leaflets, 2011, IEEE International Conference in Image Processing

[7] C.P. Wijekoon, P.H Goodwin, T.Hsiang, Quantifying fungal infection of plant leaves by digital image analysis using Scion Image software, 2008, Journal of Microbiological Methods

[8] Chaohui et al Leaf Area Measurement Based on Image Processing, International Conference on Measurement Technology and Mechatronics Automation, 2010

[9] C.P. Wijekoon, P.H. Goodwin, T. Hsiang, Quantifying fungal infection of plant leaves by digital image analysis using Scion Image software, 2008

[10] Qinghai He, Benxue Ma, Duanyang Qu, Qiang Zhang, Xinmin Hou, Jing Zhao, Cotton Pests and Diseases Detection based on Image processing, June 2013, TELKOMNIKA pp.3445~3450

[11] A.C. Nazare-JR., D. Menotti and J.M.R neves and T. Sediyma, Automatic Detection of the Damaged Leaf Area in Digital Images of Soybean, IWSSIP 2010

[12] Shi-Chia Huang, Fan-Chieh Cheng, and Yi-Sheng Chiu, 'Efficient contrast enhancement using adaptive gamma correction with weighting distribution', 2012, IEEE Transaction on Image processing, pp:99

[13] Daughtry C, Direct Measurements of Canopy Structure. Rem. Sens. Rev. 5(1):45-60

[14] Mahdi M. Ali, Ahmedi Al-Ani, Derek Eamus and Daniel K.Y. Tan, A New Image processing based Technique for Measuring Leaf Dimensions, 2012, American-Eurasian J.Agric. & Environ. Sci, pp 1588-1594

[15] Sanjay B. Patil and Shrikant K. Bodhe, Betel Leaf Area Measurement Using Image processing, 2011, IJCSE

[16] Radhakrishna Achanta, Sheila Hemami, Francisco Estrada, and Sabine Susstrunk, 'Frequency-Tuned Salient Region Detection', 2009, Computer Vision and Pattern Recognition, CVPR 2009, pp.1597-1604

[17] Jayaraman S., Veerakumar T., and Esakkirajan S. , Digital Image processing 2009, pp. 258-259

[18] Jeong-In Park and Jin-Tak Choi, 'A Background Removal Algorithm using the Variable Order n x m dimensional Vector', 2013, Proceedings, The 3rd International Conference on Circuits, Control, Communication, Electricity, Electronics, Energy, System, Signal and Simulation, 2013 (SERSC)

[19] Yiqun Hu, Xing Xie, Wei-Ying Ma, Liang-Tien Chia and Deepu Rajan, 'Salient Region Detection using Weighted Feaure Maps based on the Human Visual Attention Model.', 2005

[20] Ming-Ming Cheng, Gup-Xi Xhang, Niloy J.Mitra, Xiaolei Huang, Shi-Min Hu, 'Global Contrast based Salient Region Detection', 2011

[21] Jiang H., Wang J., Yuan Z., Liu T., Zheng N., Li S. 'Automatic Salient Object Segmentation Based on Context and Shape Prior', 2011

[22] Guanqun Cao, Faouzi Alaya Cheikh, Salient Region Detection with Opponent Color Boosting, 2010, Visual Information Processingb (EUVIP), 2rd European Workshop on, p13-p18

[23] N. Otsu, 'A threshold selection method from gray-level histogram', IEEE Transactions on System Man Cybernetics, vol. SMC-9, no.1, 1979, pp.62-66

[24] Li, Z., C. Ji and J. Liu, 2008, 'Leaf Area Calculation Based on Digital Image'. Computer and Computing Technologies in Agriculture, 259:1427-33.

**Eric Hitimana**

He received the BS degree in Computer Engineering and Information Technology from Kigali Institute of Science and Technology (KIST) Rwanda in 2010. He is graduating this coming February 2014 his MS degree in Computer Science and Engineering from Chonbuk National University, Rep. of Korea. He is an active research in Image processing

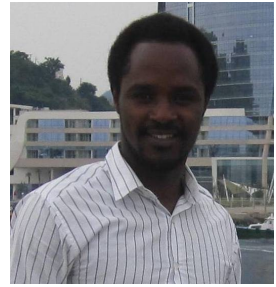

**Oubong Gwun**

He received the BS and MS degree in Electrical Engineering from Korea University in 1980, 1983 and the PhD degree in Interdisciplinary Graduate School of Engineering Sciences Kyushu University Japan in 1993. Now he is a professor of Chonbuk National University, Rep. of Korea. His interest area is Computer graphics, Image processing and Visualization.

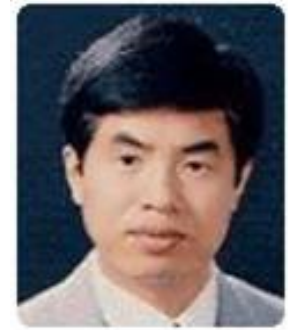